\documentclass[a4paper, 10pt, conference]{ieeeconf}      
\usepackage{FG2020}
\usepackage{multirow}
\usepackage{multicol}
\usepackage{graphicx}
\usepackage{makecell}

\FGfinalcopy 

\IEEEoverridecommandlockouts                              

\def\FGPaperID{****} 

\title{\LARGE \bf
EMOPAIN Challenge 2020: Multimodal Pain Evaluation from Facial and Bodily Expressions
}

\author{\parbox{16cm}{\centering
    {\large Joy O. Egede$^{1\dagger}$, Siyang Song$^{1\dagger}$, Temitayo A. Olugbade$^{2\dagger}$, Chongyang Wang$^{2\dagger}$, Amanda C. De C. Williams$^2$, Hongying Meng$^4$, Min Aung$^3$, Nicholas D. Lane$^5$, Michel Valstar$^1$, Nadia Bianchi-Berthouze$^{2\ast}$}\\
    {\normalsize
    $^1$ School of Computer Science, The University of Nottingham\\
    $^2$ University College London \\
    $^3$ Department of Computer Science, University of East Anglia\\
    $^4$ Department of Electronic and Computer Engineering, Brunel University London\\
    $^5$ Department of Computer Science, University of Oxford}}
    \thanks{$^{\ast}$ Corresponding author}
    \thanks{$^\dagger$ These authors made equal contributions}
}

\begin{document}

\ifFGfinal
\thispagestyle{empty}
\pagestyle{empty}
\else
\author{Anonymous FG2020 submission\\ Paper ID \FGPaperID \\}
\pagestyle{plain}
\fi
\maketitle

\begin{abstract}
The EmoPain 2020 Challenge is the first international competition aimed at creating a uniform platform for the comparison of multi-modal machine learning and multimedia processing methods of chronic pain assessment from human expressive behaviour, and also the identification of pain-related behaviours. The objective of the challenge is to promote research in the development of assistive technologies that help improve the quality of life for people with chronic pain via real-time monitoring and feedback to help manage their condition and remain physically active. The challenge also aims to encourage the use of the relatively underutilised, albeit vital bodily expression signals for automatic pain and pain-related emotion recognition. This paper presents a description of the challenge, competition guidelines, bench-marking dataset, and the baseline systems' architecture and performance on the Challenge's three sub-tasks: pain estimation from facial expressions, pain recognition from multimodal movement, and protective movement behaviour detection.

\end{abstract}

\section{INTRODUCTION}
\noindent The EmoPain 2020 Challenge \footnote{https://mvrjustid.github.io/EmoPainChallenge2020/} is the first international competition in automatic pain recognition aimed at benchmarking the performance of machine learning methods designed to recognise or quantify chronic pain from behavioural---face and body---cues, and also recognise pain-related movement behaviours. Chronic pain (CP) is a widespread distressing problem that not only restricts body activities but significantly impacts on the mental, psychological, social and economic status of people with chronic pain.  A 2016 study \cite{fayaz2016prevalence} showed that over 40\% of the UK population are affected by chronic pain with this number going up to 62\% for people over 75 years. A similar study for the United States puts the former figure at 25\% \cite{dahlhamer2018prevalence}. Beyond the individual, CP has dire consequences on socio-economic growth and development. Amongst other medical conditions, chronic pain was responsible for most medical consultations and costs the US approximately \$560 billion dollars each year \cite{dahlhamer2018prevalence}. The escalating socio-economic costs of CP, as well as its detrimental effect on the quality of life of individuals and their families, buttress the urgent need for efficient chronic pain interventions.

Technological interventions present a plausible solution, but the first step towards a workable system requires accurate identification and interpretation of pain-associated expressions and behaviours. Consequently, technology-driven methods (see survey in \cite{werner2019automatic}) utilising clinically certified behavioural and physiological pain indicators for pain assessment have been proposed within the machine learning and computer vision research community. Although machine-assisted pain assessment methods have advanced considerably, their practical application has been constrained by data-related and design issues. 

One major problem is that there are few publicly accessible pain datasets that meet requirements for effectively training such predictive systems. Secondly, pain expression is multi-faceted, yet there is an over-reliance on unimodal clues, particularly the face, whereas body movements are critical to effective chronic pain assessment \cite{Sullivan2006}. Although facial expressions give a good indication of affect intensity, without the body context, its discriminative property of affective states diminishes \cite{aviezer2012body}. In contrast, pain-related movement behaviour provides more information about the distress level of a pain stimulus (physical activity) and what form of support is required  \cite{watson1997evidence,Sullivan2006}. Thus, pain literature \cite{aung2015automatic,Sullivan2006} strongly advocates the use of multiple, rather than isolated behavioural cues for pain assessment. Lastly, existing bench-marking pain corpus \cite{walter2013biovid,lucey2012painful} predominantly feature pain expressions induced in constrained environments and by non-threatening stimuli which are not fully representative of real-world distressing physical activities encountered by people with chronic pain; whereas, for technological interventions to be beneficial, it should be developed on data which represent the everyday body functions of the target population. Also, some of these datasets \cite{lucey2012painful,brahnam2006machine} provide only uni-dimensional---facial cues---behavioural chronic pain characterisations.

The EmoPain 2020 challenge aims to address the above gaps by creating a platform to foster multi-modal automatic pain recognition research within the machine learning community. The challenge is based on the multi-modal EmoPain dataset, which for the first time, is opened up to the community in a competition framework to benchmark automated pain assessment methods. The EmoPain dataset \cite{aung2015automatic} consists of audiovisual, motion data and muscle activity captured from chronic lower back pain (CLBP) and healthy participants engaged in both instructor-led and self-directed physical exercises which replicate everyday body functions. Utilising the visual and movement data dimensions, the EmoPain 2020 challenge presents three pain recognition tasks: (i) Pain Estimation from Facial Expressions Task, (ii) Pain Recognition from Multimodal Movement Task and (iii) Multimodal Movement Behaviour Classification Task.

Participants could choose to compete in all or some of the tasks. Data for each task is split into training, validation and a held-out test partition. To ensure a fair comparison, participants were given the same training and validation data to develop their algorithms/models, which was then sent to the organisers for evaluation on the held-out test set. Participants did not have access to the test data partition. Papers accompanying the challenge submissions were presented at the FG2020 International Workshop on Automated Assessment of Pain.

The rest of the paper is organized as follows: Section II discusses relevant work in automatic pain recognition; Section III gives a full description of the EmoPain dataset and the three sub-tasks as well as the metrics used for ranking participants' submissions; Section IV describes the baseline features and models developed for each task, and the results obtained. Lastly, section V summarises the contributions and concludes the work.

\section{RELATED WORK}
\noindent This section describes current approaches to automatic pain recognition with a focus on pain-associated face and body expression synthesis, processing, analysis and interpretation. Relevant pain literature will be discussed in three groups building on the challenge's task categorisation. An extended survey is provided in \cite{werner2019automatic}.

\subsection{Automatic Pain Detection based on Facial Expressions}
\noindent The face is a key medium for communicating pain in human interactions, particularly when pain expression is not actively suppressed by the individual. Facial expressions of pain have been shown to have distinctive properties from other basic emotions \cite{kappesser2002pain,simon2008recognition}, lending credence to its pertinence to pain recognition.  Due to its relative ease of accessibility and utilisation in daily social interaction, faces have been explored extensively for automatic pain recognition. Early work based on facial actions was limited to binary classification of face images into \emph{pain} or \emph{no pain} \cite{ashraf2009painful,brahnam2007machine} or distinguishing real pain from posed pain \cite{littlewort2009automatic}. However, this outcome was not adequate for clinical applications as evidenced by the self-report pain assessment scales \cite{breivik2008assessment} which aim to quantify pain rather than identify its occurrence. Consequently, recent studies moved on to estimating pain levels from facial expressions using either a multi-class classification set-up \cite{hammal2012automatic}  or regression framework \cite{zafar2014pain,egede2017cumulative, Kaltwang_2012}. This shift was also propelled by the introduction of pain datasets \cite{walter2013biovid,lucey2012painful} which provide discrete pain annotations of face images. Most of these studies \cite{Kaltwang_2012,egede2017cumulative} predict pain on the 16-point Prckachin and Solomon Pain Intensity (PSPI) \cite{prkachin2008structure} scale or a  condensed version \cite{hammal2012automatic}, while others \cite{werner2014automatic,walter2015data} focus on recognising observer reported or patients' self-reported pain ranging from two to five pain levels.

To discriminate pain expressions, face shape and appearance descriptors have been widely employed due to their proven effectiveness in facial expressions analysis. Appearance features encode facial deformations due to expressions (e.g., wrinkles) while shape features describe the spatial localisation of facial components (i.e., eyes, mouth and nose). In terms of facial features used, previous work on pain recognition can be classified into three: (i) handcrafted feature methods \cite{werner2014automatic,hammal2012automatic,ashraf2009painful,Kaltwang_2012}, (ii) data-learned feature methods\cite{zhou2016recurrent,bellantonio2016spatio} and (iii) hybrid-feature methods \cite{egede2017cumulative, egede2019automatic}. Handcrafted facial descriptors are statistical measures computed from a face image using human-designed algorithms. Commonly used features in this category include gradients features \cite{ werner2014automatic}, Gabor features \cite{littlewort2009automatic} , Active Appearance Models (AAM) \cite{ashraf2009painful,hammal2012automatic}, Local Binary Patterns (LBP) \cite{Kaltwang_2012}, facial landmarks and associated distance metrics \cite{werner2014automatic} amongst others. Data-learned features are offshoots of neural network applications to pain recognition and are automatically generated within the network. Hybrid features, on the other hand, are an integration of traditional and data-learned features and have been shown to significantly improve the predictive ability of recognition models on small datasets \cite{egede2017fusing}.  

Although pain recognition from faces has witnessed tremendous progress, there is still ample scope for improvement. Current work has concentrated on facial data collected in constrained, ideal settings where several video cameras are positioned at strategic positions to capture face images from all possible angles. Thus, captured images are usually high resolution, near frontal and unobstructed faces, whereas this is not always the case in typical everyday settings, e.g., performing rehabilitation exercise at home. Another open challenge is insufficient data representation for higher pain levels in existing pain corpus, which limits the performance of recognition models on these pain classes \cite{egede2017fusing}. Hence, novel methods that make the most of existing data, and more focus on the creation of representative chronic pain facial data are required.

\subsection{Automatic Pain Detection based on Bodily Expressions}
\noindent Despite findings in \cite{Sullivan2006} that the body may be more expressive of pain experience than the face or vocal modality, which are more dependent on social context, it has not been as widely explored for automatic detection of pain levels as the face. Most of the early studies \cite{Ahern1988,Gioftsos1996} and a number of more recent work \cite{Grip2003,Lai2009} focused on discrimination between people with chronic pain and those without. Other studies have similarly investigated differentiation between two levels of pain \cite{Bishop1997,Rivas2015}. One exception is \cite{Dickey2002} where  11 levels of pain were detected. While studies such as \cite{Kachele2016, Werner2016} have also gone beyond binary classification, unlike the afore-mentioned, they are based on experimentally-induced pain which is transient and not usually perceived a threat \cite{Legrain2009}.

The bodily expressions used in the investigations carried out in these studies have typically depended on the pain location and the activity being performed. For example, in the work of \cite{Lai2009}, automatic detection of knee pain was based on gait characteristics and ground force reaction during walking tasks. Similarly, the automatic detection of neck pain in \cite{Grip2003} used neck movements measured while participants performed neck exercises. For low back pain, where participants are usually being assessed during physical activities involving the trunk, features of trunk \cite{Ahern1988,Gioftsos1996,Bishop1997}, spine \cite{Dickey2002}, knee \cite{Gioftsos1996}, and hip \cite{Gioftsos1996} movement, corresponding back muscle activity, and force and centre of gravity \cite{Gioftsos1996} have been used for pain (level) detection.

Another work in the area related to body movement is the one of Rivas et al. \cite{Rivas2018}. In their work, the authors explore the use of hand pressure and joystick manipulation to detect stroke patients' pain level by personalising the model to each patient by using data from 10 different sessions. In \cite{Rivas2019}, the authors extend the work by combining multiple modalities (hand pressure, gesture and facial expressions) to investigate the relationship between affective states and pain during rehabilitation. Again, individual models are built by taking advantage of the multiple sessions. 

In a recent study \cite{Olugbade2019} on automatic discrimination between healthy participants, low-level pain, and high-level pain based on complete movement instances in the EmoPain dataset, we explored features of the trunk, knee, head/neck, and arm movements computed from full-body positional data as well as features from shoulder and lower back muscle activity. We used two separate sets of features for trunk flexion and sit-to-stand movements respectively, given the considerable differences in the temporality of the two movements and the anatomical regions recruited in performing them. We additionally built a separate model for each movement type for this reason and especially to manage the limited data size available. For full and forward trunk flexion, we extracted the range of trunk and neck movement, the amount of unsteadiness in arm movement, and the time and amplitude of high-to-low muscle activity change; for sit-to-stand, we extracted range of trunk and neck movement, knee and pelvic angles at the point of buttocks lift, speed and duration of the lift phase, and the time of high-to-low muscle activity change and muscle activity range. We obtained 0.90 F1 score (0.90 accuracy) on average, over the three classes and three movement types, based on leave-one-subject-out cross-validation. 

\subsection{Automatic Detection of Protective Movement Behaviour}
\noindent Aside from the pain estimation on bodily expressions, the movement behaviour presented therein is informative not only of pain level but also of the emotional state and engagement level of people with chronic lower back pain (CLBP). Specifically, the protective behaviour, e.g., hesitation, guarding, stiffness, the use of support and bracing \cite{Keefe1982development}, expression of fear or low-efficacy of movements, is currently adopted by physiotherapists in tailoring their feedback and interventions \cite{Vlaeyen2000, Olugbade2019relationship}. As the rehabilitation for CLBP people is moving towards self-management outside the hospital, researchers started to work on the establishment of a virtual physiotherapist, where the first step is about the automatic detection of protective behaviour. Early studies in this direction mainly focused on feature-engineering methods to extract discriminative features from motion capture (MoCap) and surface electromyographic (sEMG) data with shallow classifiers like Random Forests and Support Vector Machine applied on top of them \cite{aung2015automatic,OlugbadeSelfEfficacy18,Aung2014behavior}. To name a few, features used include the range of joint angle, the mean of the angular velocity and the mean of the upper-envelope of the sEMG data. One limitation of these works is the lack of generalisability across different types of movement. Recently, efforts are also seen in using deep learning for the detection of protective behaviour. A comparison of different vanilla neural networks is provided in \cite{wang2019recurrent}, while some data augmentation techniques were also explored. The result achieved is much higher than previous feature-based methods, on the data pooled from different movement types. Later on, a collaboration of LSTM network with attention mechanism is presented in \cite{Wang2019attention}, where better and explainable results are reported. However, challenges still exist, such as the dependence on the pre-segmented activity sequences which is not able to provide real-time encouragements and feedback, and the lack of exploitation of the bio-mechanical nature of MoCap and sEMG data especially, resulting from the traversal data processing strategy.

\section{Challenge Description}
\noindent This section describes the data collection protocol for the benchmark data (EmoPain database), the Challenge's tasks, task data partitioning, and proposes real-world applications of each task to clinical pain management.

\subsection{EmoPain Dataset}
\noindent The EmoPain dataset \cite{aung2015automatic} provided for the challenge originally comprised of audiovisual, motion-capture and muscle activity data, collected from 18 CLBP and 22 healthy participants. Here need to note that, the real number of participants provided for each challenge task differs. Each participant went through at least one trial of the data collection, either the normal or the difficult trial. Within a trial, the participant performs a sequence of activities, namely one-leg-stand, reach-forward, stand-to-sit, sit-to-stand and bend-down. These activities are connected by transition activities, like standing still, sitting still and self-preparation. In the difficult trial, participant has to follow instructions set by the experimenter and carry a 2Kg weight in each hand during the performance of reach-forward and bend-down. There are no such limitations in the normal trial. 

For the facial expression video, several sets of features are extracted for the challenge participant, which will be described in detail in the next section. For the body movement data, the joint angles and respective angular velocities are computed. The dataset for the challenge is split into training, validation and a held-out test partition. The participant partition are shown in Table \ref{tab:data_partitioning}. The class distribution is not considered for the partition of the dataset, but we ensure each partition has sufficient representation of healthy participants and CLBP patients' data.

\subsection{Challenge Tasks}
\noindent The EmoPain Challenge consists of three main tasks namely: (i) pain estimation from facial expressions, (ii) pain recognition from multi-modal movement, and (iii) protective movement behaviour detection. Participants were expected to compete in at least one or more tasks. 

The \textbf{\emph{Pain Estimation from Facial Expressions Task}} aims to develop technology to automatically quantify pain from face images of CLBP and healthy participants performing physical activity. These technologies could potentially support real-time pain assessment for patients who are unable to self-report pain, e.g., unconscious patients, and in constrained settings, e.g.,  ICUs, where continuous recording of a person's face is possible.
Anchoring on facial properties deemed suitable for facial expression analysis \cite{Kaltwang_2012,egede2017fusing}, data for this sub-task consists of anonymized face shape and appearance features extracted from the EmoPain video images (see details in \ref{baseline_features_face}), as well as observer pain annotations for each face image on an 11-point scale ranging from 0 (no pain) to 10 (maximum possible pain intensity).Due to data protection and ethical constraints, we did not provide the original video images. 

Note that the values of the original pain annotations for the face range from $0$ to $1000$. These labels are heavily unbalanced, as the value of most labels are zero and for some other values, only less than $10$ frames have such pain level. To alleviate this problem, we re-sampled all labels into $11$ bins, from $0$ to $10$. Specifically, the values of all original labels were divided by $100$, and then allocated to the bin whose value corresponds to their integral part, e.g., a label value of 232 will be assigned to \textit{bin $2$}. The distribution of the final provided labels are detailed in Table \ref{tab:facial_label}. 
Participants' submissions to this task were ranked using the \emph{Concordance Correlation Coefficient (CCC)} \cite{lawrence1989concordance} which measures the temporal association between the model predictions and ground truth pain labels. CCC is preferred over similar measures--- \emph{Pearson's CC} and \emph{Spearman's CC}--- because it encodes precision and accuracy metrics in a single measurement and is robust to location and scale variations \cite{lawrence1989concordance}.

The \textbf{\emph{Pain Recognition from Multimodal Movement Task}} aims to detect and classify levels of pain experienced by a person with chronic pain during movement activities. Technology with this capability could help a person with chronic pain more helpfully pace physical activity performance \cite{Olugbade2019}. Data for this sub-task comprises of muscle activity data, 13 joint angles and angular energies (see full description in \cite{Wang2019attention}) captured from CLBP  and healthy participants while performing physical activities. Each activity instance is accompanied by a three-class pain annotation: no pain, low pain and high pain, which will serve as ground-truth labels for the task. The submissions for this task were evaluated using F1 scores and accuracy, but final ranking was done based on Matthew Correlation Coefficient (MCC) \cite{Matthews1975} which better accounts for the negative classes.  

The \textbf{\emph{Multimodal Movement Behaviour Classification Task}} aims to develop technology that can detect and classify protective behaviours (e.g., rigid movement) in people with chronic pain. Such technologies could provide immediate and appropriate feedback or support to users, e.g., notifying the user to adopt a correct posture if the use of maladaptive strategy is detected \cite{Olugbade2019, wang2019recurrent}. Data for this task consists of 13 bodily joint angular features and muscle activity for each movement frame with corresponding activity-type labels and binary protective behaviour annotations by 2 physiotherapists and 2 psychologists. For this task, macro average F1 score and the F1 score for each class (i.e. protective and non-protective) were used for ranking participants' submissions.

\begin{table}[!t]
    \caption{Participant distribution in each data partition. CLBP - Chronic Lower Back Pain; HP - Healthy Participants}
    \label{tab:data_partitioning}
    \centering
    \resizebox{1\linewidth}{!}{ 
     \begin{tabular}{|l|l|l|}
            \hline
            Partitions & Face Tasks & Body Tasks\\
            & &   \\ \hline
            Train & 8 CLBP and 11 HP& 10 CLBP and 6 HP  \\ \hline
            Validation & 3 CLBP and 6 HP & 4 CLBP and 3 HP  \\ \hline
            Test & 3 CLBP and 5 HP & 4 CLBP and 3 HP \\ \hline      
        \end{tabular}
    }
\end{table}


\begin{table*}[!t]
    \caption{Label distribution of the pain estimation from face sub-challenge }
    \label{tab:facial_label}
    \centering
    \resizebox{1\linewidth}{!}{ 
    \begin{tabular}{|c|c|c|c|c|c|c|c|c|c|c|c|}
        \hline
        Label value & 0 & 1 & 2 & 3 & 4 & 5 & 6 & 7 & 8 & 9 & 10\\  \hline
        Training & 646634 & 39694 & 31032 & 61148 & 41286 & 17122 & 16958 & 9140 & 3734 & 626 & 2078\\ 
        Development & 475717 & 20731 & 31697 & 25613 & 20765 & 15416 & 7425 & 9972 & 198 & 176 & 218\\ \hline        
    \end{tabular}
    }
\end{table*}
\section{BASELINE FEATURES AND SYSTEMS}
\noindent In this section, we describe the features extracted from each pain expression modality, the baseline models implemented for each task, and present the results obtained from the performance evaluation of the models.

\subsection{Pain Estimation from Facial Expressions}
\label{baseline_features_face}
\noindent For the pain estimation from face sub-challenge, we extracted four facial descriptors using the OpenFace 2.0 toolkit \cite{baltrusaitis2018openface}, and two deep-learned emotion-oriented feature representations \cite{ringeval2019avec}. The  detailed descriptions of these features are as follows:
\begin{itemize}
    \item \textit{Facial landmarks}: 68 2-D and 3-D fiducial facial points.
    \item \emph{Head pose}: Pitch, yaw and roll angles.
    \item \emph{Gaze}: 3-D gaze directions.    
    \item \emph{HOG}: a $4464$-D Histogram of Oriented Gradients (HOG) features. 
    \item \emph{Action Unit (AU) occurrence}: $18$ AUs whose values are $1$ (present) or $0$ (absent).
    \item \emph{AU intensities}: $17$ AUs whose values range from 0 to 5 (max intensity).
    \item \emph{VGG-16 feature}: $4096$-D deep features extracted from the second fully-connected layers of the VGG-16 network \cite{simonyan2014very}. 
    \item \emph{ResNet-50 feature}: $2048$-D deep features extracted from the fully-connected layers of the ResNet-50 network. \cite{he2016deep}. The VGG-16 and ResNet-50 network are pre-trained on the Affwild dataset \cite{kollias2019deep} with valence and arousal labels.
\end{itemize}

\noindent Although the data labels are significantly imbalanced as seen in Table \ref{tab:facial_label}, we do not  perform any data augmentation, to enhance the reproducibility of the reported results. While the task can be solved as an 11-class classification problem, in this challenge, we treated it as a regression problem.

The face baseline system employed four different feature sets:  $2$ hand-crafted features including geometric features (a combination of 2-D facial landmarks and gaze directions) and $4464$-D HOG feature; and $2$ emotion-oriented deep-learned feature sets including $4096$-D VGG-16 features and $2048$-D ResNet features. Note that the 2-D facial landmarks are transformed into a $136$-D dimension feature vector for each frame. The training process starts with feature normalisation. For each dimension of the input feature, the training set was normalised using z-score as shown in Equation \ref{z_score}.
\begin{equation}
    \label{z_score}
    z = \frac{x-\mu}{\sigma}
\end{equation}
where $\mu$ and $\sigma$ are the mean and standard deviation of the feature values over the entire training data. The obtained mean value and standard deviation were then applied to normalize the validation and test set. In this sub-challenge, we trained an Artificial Neural Network (ANN) for each feature subset. The employed ANNs follow the set-up presented in \cite{Jaiswal2019automatic}, which consists of $4$ fully connected hidden layers. A dropout \cite{srivastava2014dropout} with probability $0.5$ and a ReLU layer is placed after each fully-connected layer. RMSprop is used as the training method, while Mean Square Error (MSE) is employed as the loss function. The hyper-parameters and topology chosen for the baseline systems are shown in Table \ref{tab:facial_hyper}. These hyper-parameters were determined by grid search on validation set.

The baseline results of the Pain Estimation from Faces sub-challenge are given in Table \ref{tab:face_baseline_result}. They show that amongst the single-feature models, the best correlation (CCC) on the development set results was achieved by VGG-16 feature, which also obtained good RMSE and MSE results. However, while VGG-16 feature also achieved solid performance on the test set in terms of the RMSE and MSE, its predictions are not highly correlated with the ground-truth of the test set. Instead, the combination of facial landmarks and eye gaze features produced excellent RMSE and MSE results on both development and test set, and also generated predictions with the highest correlation (PCC) to the labels in the test set. These results indicate that the pain level can be partially reflected by the geometric information of the face and eyes. 

The decision-level fusion of all modalities gave the best results on both the development set (RMSE $=1.69$, PCC $=0.25$, CCC $=0.18$) and test set (MAE  $=0.91$, RMSE $=1.41$, PCC $=0.10$, CCC $=0.06$), except the MAE returned on the development set (MAE $=1.26$) is slightly higher than the best one (MAE $=1.24$). Based on the fusion results, we can argue that though the individual features were not very informative for pain intensity estimation when simple ANNs are used as the back-end, their fusion still seems to provide more valuable and \textbf{positive} information for pain estimation. Based on all results, the recognition of pain intensities from the face is still challenging when only combining existing standard hand-crafted or deep-learned features with a simple back-end. This observation opens interesting research questions about how to extract pain-related cues from complex facial expressions and emotions. 

\begin{table}[b]
    \caption{The chosen hyper-parameters of ANNs for facial challenge baseline systems}
    \label{tab:facial_hyper}
    \centering
    \resizebox{1\linewidth}{!}{ 
    \begin{tabular}{|l|c|c|c|}
        \hline
        Feature & Hidden Layers Size & Learning Rate & Batch Size\\ \hline
        FL+Gaze & (128, 64, 32, 32) & 0.001 & 128\\
        HOG & (2000, 512, 256, 64) & 0.001 & 256\\
        VGG-16  & (1024, 256, 64, 64) & 0.005 & 128\\ 
        ResNet-50 & (1024, 256, 64, 64) & 0.001 & 256\\ \hline
    \end{tabular}
    }
\end{table}

\begin{table}[!b]
    \caption{Baseline results for the Pain Estimation from Facial features. Best results are highlighted in bold}
    \label{tab:face_baseline_result}
    \centering
    \resizebox{1\linewidth}{!}{ 
    \begin{tabular}{|l|c|c|c|c|c|}
        \hline
        Modality & Partition & MAE & RMSE &PCC &CCC\\ \hline
        FL+GAZE & Valid.& 1.51 & 1.74 & 0.04 & 0.003\\
        FL+GAZE & Test.& 1.37 & 1.56 & \textbf{0.10} & 0.003\\        
        HOG& Valid.&\textbf{1.24} & 1.91 & 0.05 & 0.04\\
        HOG& Test.& 0.93 & 1.61 & 0.03 & 0.02\\        
        VGG-16& Valid.&1.34 & 1.82 & 0.24 & \textbf{0.18}\\
        VGG-16& Test.& 0.92 & 1.43 & 0.02 & 0.004\\        
        ResNet-50& Valid.&1.42 & 2.08 & -0.08 & -0.04\\
        ResNet-50& Test.& 1.14 & 1.74 & -0.09 & -0.06\\ \hline 
        Fusion& Valid.&1.26 & \textbf{1.69} & \textbf{0.25}& \textbf{0.18}\\  
        Fusion& Test.& \textbf{0.91} & \textbf{1.41} & \textbf{0.10} & \textbf{0.06}\\  \hline       
    \end{tabular}
    }
\end{table}

\subsection{Pain Classification based on Body Movement and Muscle Activity}
\noindent Due to the limited data size available in this task, we chose to build a single model for all movement types in the dataset so as to maximise the training data. The features that we extracted (see Table \ref{tab:painmovtfeats}) were based on findings in \cite{Olugbade2019}. We extracted range of joint angles, to characterise the range of movement across anatomical regions relevant to the movement types. We additionally computed speed of movement over all joint angles and over each movement. While it might ordinarily be valuable to compute speed separately for each joint, it was necessary for us to constrain feature dimensionality in order to further address the data size limitation. Finally, we computed the range of activity for each of the four muscle groups in the sEMG data. 

Each data instance is made up of one or more iterations (up to 6) of a complete movement type, and so it was important to additionally incorporate the dynamics within each instance in the feature set. We addressed this by extracting the 18 above-mentioned features in 4 identically-sized non-overlapping window segments that together cover the data instance. 4 was a compromise between limiting the number of features and characterising movements which had the maximum number of repetitions. This led to 72 dimensions for the feature vector for each data instance.

We explored three main algorithms for the three-level classification of pain based on body movement and muscle activity data: Random Forest (RF) \cite{Breiman2001}, Support Vector Machines (SVMs) \cite{Cortes1995}, and k-Nearest Neighbours (kNN). The algorithms were evaluated using leave-one-subject-out cross-validation, based on the challenge training set alone. The hyperparameters for the algorithms were set based on grid search using an inner validation set within each validation fold, and among: 1, 5, 10, and 50 trees for the RF, and one, square root of the total amount, and the total amount for the number of features used to split each node in the RF; 1 to 5 degrees for the polynomial SVM, Gaussian or sigmoid kernels for the SVM, and 0.001, 0.01, 0.1, 1, 10, and 100 as the box constraint size for either of the three SVMs; k between 1 and 5, and \textit{minkowski}, \textit{euclidean}, \textit{manhattan}, or \textit{chebyshev} distances for the kNN. Note that in the SVMs and kNN setup, the feature set was normalised to zero and unit variance.

\begin{table}[t]
    \caption{Bodily Features used for Pain Classification}
    \label{tab:painmovtfeats}
    \setlength{\tabcolsep}{3.5pt}
    \renewcommand{\arraystretch}{1.5}
    \centering
    \begin{tabular}{|l|c|c|c|}
        \hline
        \multicolumn{2}{|c|}{Features} & Formulae & Dimension \\ \hline
        \multicolumn{2}{|l|}{Range of joint angle} & $\Delta{J_i}=\max_t J_i - \min_t J_i$ & 11 \\ \hline
        \multirow{3}{*}{Speed} & max & $\max_i \max_t \frac{\delta J_i}{\delta t}$ & 1 \\ \cline{2-4}
        & min & $\min_i \min_t \frac{\delta J_i}{\delta t}$ & 1  \\ \cline{2-4} 
        & mean & $\frac{\sum_i \frac{\sum_t \frac{\delta J_i}{\delta t}}{T}}{I}$ & 1 \\ \hline
        \multicolumn{2}{|l|}{Range of muscle activity} & $\Delta{E_k}=\max_t E_k - \min_t E_k$ & 4\\ \hline
        \multicolumn{4}{|l|}{where i = 2, 3, ..., I; I = 13; t = 1, 2, ..., T; k = 1, 2, 3, 4} \\ \hline
    \end{tabular}
    
\end{table}

The kNN, and sigmoid and Gaussian SVM, which emerged as not worse off than chance-level detection based on the cross-validation, were further evaluated in hold-out validation, with the challenge training, validation, and test sets for training, validation, and testing respectively. Table \ref{tab:painmovtdatasizes} shows the data sizes across the three pain classes (healthy, low-level pain, and high-level pain) for both the leave-one-subject-out cross-validation (LOSO-CV) and the hold-out validation. Table \ref{tab:painmovtresultslosocv} shows the F1 scores, Matthews Correlation Coefficients (MCCs) \cite{Matthews1975}, and accuracies of the SVM, RF, and kNN, for three-level pain classification based on leave-one-subject-out cross-validation with the training set. Both the RF and polynomial SVM perform worse than chance-level detection (F1 score = 0.33; MCC = 0; accuracy = 0.33). As can be seen in Table \ref{tab:painmovtresultsholdout}, although the non-polynomial SVM has the best performance in the cross-validation, it performs much poorly in further evaluation on the test set, whereas the kNN has more or less the same performance in both the cross-validation and the hold-out validation, albeit only about as good as chance-level detection. In the cross-validation, the kNN performs worst in detection of the high-level pain class (F1 score = 0.16, MCC = -0.02) compared with the healthy class (F1 score = 0.44, MCC = 0.1) and the low-level pain class (F1 score = 0.41, MCC = 0.08). However, in hold-out validation, its performance is worst for the low-level pain class (see Table \ref{tab:painmovtresultsholdout}).

\begin{table}[t]
    \caption{Data Sizes for MoCap and sEMG Data for Pain Classification}
    \label{tab:painmovtdatasizes}
    \setlength{\tabcolsep}{3.5pt}
    \renewcommand{\arraystretch}{1.2}
    \centering
    \begin{tabular}{|l|c|c|c|}
        \hline
        \multirow{2}{*}{Pain Class} & 
        \multirow{2}{*}{\makecell{Training Set}} & 
        \multirow{2}{*}{\makecell{Validation Set}} & 
        \multirow{2}{*}{\makecell{Test Set}} \\
        & & & \\ \hline
        Healthy & 34 & 25 & 25 \\
        Low-Pain & 44 & 30 & 4 \\
        High-Pain & 35 & 5 & 26 \\ \hline
    \end{tabular}
\end{table}
\begin{table}[t]
    \caption{LOSO-CV  Baseline Results for Pain Classification from MoCap and sEMG Data}
    \label{tab:painmovtresultslosocv}
    \setlength{\tabcolsep}{3.5pt}
    \renewcommand{\arraystretch}{1.2}
    \centering
    \begin{tabular}{|l|c|c|c|}
        \hline
        \multirow{2}{*}{Algorithm} & \multirow{2}{*}{\makecell{F1 Score*}} & \multirow{2}{*}{MCC*} & \multirow{2}{*}{Accuracy} \\ 
        & & &  \\ \hline
        Sigmoid/Gaussian SVM & 0.41 & 0.19 & 0.44 \\ \hline
        kNN & 0.34 & 0.05 & 0.37 \\ \hline
        RF & 0.26 & -0.10 & 0.27 \\ \hline
        Polynomial SVM & 0.15 & -0.16 & 0.26 \\ \hline
    \end{tabular}
\end{table}

\subsection{Protective Movement Behaviour Detection}
\noindent To leave enough space for explorations, a stacked-LSTM network adapted from \cite{wang2019recurrent} is used as the baseline for the movement behaviour detection task. The architecture stays the same, where three LSTM layers with 32 hidden units are used together with a softmax fully-connected layer for classification. The input to the network is a frame with size of NxTxD, where N is the number of samples, T is the length of timesteps and D is the dimension of features. The data used is the 13 angles and their respective square of angular velocities as well as the upper envelope of the sEMG data. As a result, the data matrix has the dimension D=30. A sliding window of 180 timesteps long and a 0.75 overlapping ratio is used to extract consecutive frames from each activity type. To enable the training of stacked-LSTM, we further applied two augmentations: i) jittering, where Gaussian noise with standard deviation of 0.05, 0.1 and 0.15 are globally applied to the raw data; ii) cropping, where samples at random timesteps and body parts are set to 0 with probability of 0.05, 0.1 and 0.15. Augmentation is only applied to training data. The number of frames after segmentation is 6623 (with protective frames totalling 1,330), which is augmented to 33,115 (with protective frames totalling 6,650). The hold-out validation stays the same with the other two tasks. The groundtruth of each frame is determined by majority-voting: a frame is labelled as protective if at least half of the samples within it were coded as protective, and vice versa.

The results achieved by the stacked-LSTM network are reported in Table \ref{tab:protectiveresults}. We can see from the result that all the frames in the validation set are detected as non-protective. This can be due to the fact that the protective and non-protective samples included in the training set are very imbalanced, while the baseline method does not apply any technique to solve it. On the other hand, the size of the training data is still limited. The result on the test set is slightly better with some frames correctly detected as protective (F1 score of protective class=0.2465). This proved the feasibility of using deep learning for the detection of protective behavior. Except for processing the MoCap and sEMG in a traversal way that ignored the biomechanical connectivity, challenges remain on i) how to deal with the imbalance problem in the data set; ii) how to design better data augmentation approaches. 

\begin{table}[t]
    \caption{Hold-Out Validation Baseline Results for Pain Classification using MoCap and sEMG Data}
    \label{tab:painmovtresultsholdout}
    \setlength{\tabcolsep}{3.5pt}
    \renewcommand{\arraystretch}{1.2}
    \centering
    \begin{tabular}{|l|c|c|c|c|}
        \hline
        
        \multirow{3}{*}{} & \multicolumn{2}{c|}{\multirow{3}{*}{\makecell{kNN \\ (k=1, manhattan distance)}}} & \multicolumn{2}{c|}{\multirow{3}{*}{\makecell{Sigmoid/Gaussian SVM \\ (Gaussian kernel, \\ box constraint=0.1)}}}\\ 
        & \multicolumn{2}{c|}{} &\multicolumn{2}{c|}{} \\
        & \multicolumn{2}{c|}{} &\multicolumn{2}{c|}{} \\ \hline
        \multirow{2}{*}{Metric} & \multirow{2}{*}{F1 Score} & \multirow{2}{*}{MCC} & \multirow{2}{*}{F1 Score} & \multirow{2}{*}{MCC} \\
        & & & &   \\ \hline
        Healthy (0) & 0.39 & -0.04 & 0.00 & - \\
        Low-Pain (1) & 0.09 & -0.06 & 0.14 & -\\
        High-Pain (2) & 0.44 & 0.16  & 0.00 & - \\ \hline
        Average & 0.31 & 0.02 & 0.34 & -  \\ \hline
        & \multicolumn{2}{c|}{} & \multicolumn{2}{c|}{}  \\ \hline
       Accuracy & \multicolumn{2}{c|}{0.35} & \multicolumn{2}{c|}{0.07} \\ \hline
     
    \end{tabular}
\end{table}

\begin{table}[t]
\caption{Baseline Hold-out Validation Results for Protective Movement Behaviour Detection with MoCap and sEMG data}
    \label{tab:protectiveresults}
    \setlength{\tabcolsep}{3.5pt}
    \renewcommand{\arraystretch}{1.2}
    \centering
    \begin{tabular}{|c|c|c|c|c|}
\hline
Method                        & Partition              & Class              & Acc    & F1 score \\ \hline
\multirow{6}{*}{stacked-LSTM} & \multirow{3}{*}{Valid} & Non-protective (0) & -      & 0.9622   \\ \cline{3-5} 
                              &                        & Protective (1)     & -      & -        \\ \cline{3-5} 
                              &                        & Average            & 0.4636 & 0.4811   \\ \cline{2-5} 
                              & \multirow{3}{*}{Test}  & Non-protective (0) & -       & 0.9029   \\ \cline{3-5} 
                              &                        & Protective (1)     & -       & 0.2465   \\ \cline{3-5} 
                              &                        & Average            & 0.828  & 0.5747   \\ \hline
    \end{tabular}
\end{table}


\section{CONCLUSION}
\noindent In this paper, we introduced the first EmoPain 2020 Challenge on automatic pain recognition from multimodal face and body expressions based on the EMOPAIN dataset and guidelines for participation in the competition. It featured three tasks: (i) pain estimation from face shape and appearance features, (ii) pain recognition from muscle activity and joint angle statistical features, and (iii) classification of protective body movement behaviour. 
For each task, we described the expressive behavioural features extracted, the baseline system implementations and performance on the benchmark dataset. In this challenge, participants only received the extracted expression features rather than the video data, thus the baseline implementations do not employ feature optimisation or augmentation methods to allow for reproducibility of the results.  Lastly, the baseline program code, results and participant rankings can be found on the EmoPain2020 Challenge's webpage: 	\emph{https://mvrjustid.github.io/EmoPainChallenge2020/}.
\section{ACKNOWLEDGMENTS}
This work was funded by the EPSRC grant Emotion \& Pain Project EP/H017178/1 and the NIHR Nottingham Biomedical Research Centre.




\bibliographystyle{IEEEtran}
\bibliography{main}

\end{document}